\documentclass[letterpaper, 10 pt, journal, twoside]{ieeetran}

\IEEEoverridecommandlockouts                              

\usepackage{graphics} 
\usepackage{algorithm}
\usepackage{algorithmic}
 \usepackage{dblfloatfix}
\usepackage{amsfonts,amssymb,amsmath,bm}
\usepackage{enumerate}
\usepackage{cite}
\usepackage[normalem]{ulem}
\usepackage{caption}
\usepackage{adjustbox}
\usepackage{booktabs}
\usepackage{multirow}
\usepackage{csquotes}
\makeatletter
\newcommand\notsotiny{\@setfontsize\notsotiny\@vipt\@viipt}
\makeatother
\usepackage{xcolor}
\usepackage{subcaption}
\usepackage{rotating}  

\usepackage[font=footnotesize,labelfont=bf,tableposition=top]{caption}
\usepackage{diagbox}
\usepackage{hyperref}
\hypersetup{
    colorlinks,
    linkcolor={red!50!black},
    citecolor={blue!50!black},
    urlcolor={blue!80!black}
}

\title{
Semantic–Geometric Task Representations for Bimanual Manipulation from Human Demonstrations to Robot Action Planning
}

\author{Franziska Herbert$^{1, 2}$, Vignesh Prasad$^{1, 2, 3}$, Han Liu$^{1, 2}$, Dorothea Koert$^{1}$ and Georgia Chalvatzaki$^{1, 2, 3}$%
\thanks{Manuscript received: December, 29, 2025; Revised April, 23, 2026; Accepted May, 20, 2026. This paper was recommended for publication by Editor Júlia Borràs Sol upon evaluation of the Associate Editor and Reviewers’ comments. Contact: {\tt\small franziska.herbert@tu-darmstadt.de}} 
\thanks{$^{1}$ Department of Computer Science, TU Darmstadt, Germany. $^{2}$ Hessian.AI. $^{3}$ Robotics Institute Germany (RIG). }%
\thanks{This research is funded by EU Horizon Europe Projects MANiBOT (101120823), ARISE (101135959), the DFG Emmy Noether Programme (CH 2676/1-1), the ERC project \enquote{SIREN} (101163933) and the German Federal Ministry of Research, Technology and Space (BMFTR) Projects \enquote{RIG} (16ME1001) and \enquote{IKIDA} (01IS20045). }
\thanks{The authors gratefully acknowledge the computing time provided on the high-performance computer Lichtenberg II at TU Darmstadt, funded by the BMFTR and State of Hesse. }
\thanks{Digital Object Identifier (DOI): 10.1109/LRA.2026.3703251}
}

\let\oldtwocolumn\twocolumn
\renewcommand\twocolumn[1][]{%
    \oldtwocolumn[{#1}{
    \begin{center}
           \includegraphics[width=1\textwidth]{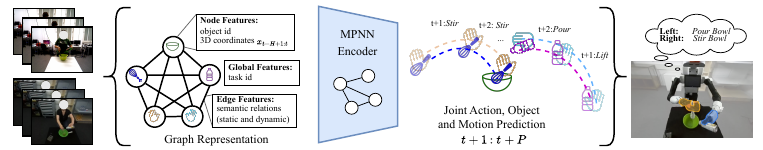}
           \captionof{figure}{Overview of our approach. We learn graph representations from bimanual human demonstrations of manipulation tasks. A Message Passing Neural Network is used to learn the underlying concepts of the task, which can subsequently be used to replicate the task on other agents or environments.
           }
           \label{img:pipeline}
        \end{center}
    }]
}

\begin{document}

\maketitle

\begin{abstract}
Learning structured task representations from human demonstrations is essential for bimanual manipulation, where action ordering, object involvement, and interaction geometry vary significantly across executions. A key challenge lies in jointly capturing the discrete semantic task structure and the temporal evolution of object-centric geometric relations in a form that supports reasoning over task progression. We introduce a semantic--geometric graph-based task representation that jointly encodes object identities, inter-object semantic relations, and per-object motion histories, via a Message Passing Neural Network (MPNN) encoder and a Transformer-based decoder. The encoder operates solely on the temporal scene graph, producing structured representations decoupled from action labels. The decoder then conditions on action-context to forecast future actions, associated objects, and object motions. This decoupling learns task-agnostic representations, enabling encoder reuse across embodiments through decoder-only finetuning on a small robot dataset. Across eleven bimanual tasks from two datasets, we find that the benefit of structured semantic--geometric representations over simpler sequence-based models grows with task variability in action ordering and object involvement. At deployment, a planner couples the action and motion predictions with learned Probabilistic Movement Primitives, achieving full task success on two real-robot bimanual tasks and outperforming graph ablations, Transformer, decoder-only, and finetuned vision-language model baselines. Website: \url{https://frherbert.github.io/bimanual-task-graphs}
\end{abstract}


\section{Introduction}

\IEEEPARstart{B}{imanual} manipulation tasks exhibit substantial variability in their executions. 
Consider clearing a cluttered table into a box: the order in which objects are selected, which hand grasps each item, and the geometry of each interaction all vary across demonstrations. Learning generalizable structured task representations across such variability requires capturing both \textit{what} is happening (the semantic structure of actions, objects, and their interactions) and \textit{how} it happens (the geometric evolution of the scene over time). Without both, a learned representation either memorizes sequences it cannot generalize from, or tracks motion without understanding task structure.

Scene Graphs~\cite{armeni20193d,gay2019visual} capture semantic relationships and geometric information in a structured representation, making them a natural choice for manipulation tasks, where reasoning about task progression requires both discrete action sequences and continuous geometric evolution. Graph Neural Networks~\cite{grinberg2023introduction, wu2020comprehensive, bronstein2021geometric} aggregate this information across the graph through message passing. However, existing scene-graph-based approaches to task understanding~\cite{lagamtzis2023graph, razali2023action, dreher2019learning} typically emphasize either semantic structure (e.g., action--object relations) or geometric evolution (e.g., motion trajectories), but rarely integrate both in a unified manner. As a result, these methods remain at the offline prediction level of frame-wise action recognition~\cite{dreher2019learning}, motion forecasting~\cite{razali2023action}, or single-step unimanual predictions~\cite{lagamtzis2023graph}, without showing their ability to drive robotic decision-making.

We argue that learning task representations for bimanual manipulation requires both joint modeling of semantic relations and geometric evolution, and a forecasting scope that reaches beyond single-frame or single-step predictions. We show that representations meeting these requirements can serve as the substrate for online robot action planning. To this end, we introduce a semantic--geometric task graph-representation and a GNN-based encoder that jointly encodes object identities, inter-object semantic relations, per-object motion histories, and global task context into coupled node, edge, and global embeddings, in contrast to prior graph-based approaches that encode only a subset of these.

Crucially, we learn task-level representations that capture both the semantic action structure and the evolution of geometric scenes. To this end, our decoder predicts future action sequences together with their associated objects and individual object motions from a shared graph-embedding bottleneck, enabling joint forecasting of symbolic task progression and continuous object motion. \textit{At deployment}, a planner scores candidate robot skills by combining predicted action likelihoods with motion-consistency under learned Probabilistic Movement Primitives~\cite{paraschos2013probabilistic}, exploiting the semantic and geometric prediction heads at test time.

Our key contributions are:
\textbf{(1)} a semantic--geometric task graph-representation that jointly encodes object identities, inter-object relations, and per-object motion histories within a single graph for bimanual manipulation from human demonstrations;
\textbf{(2)} a decoupled graph-encoder/transformer-decoder architecture that separates scene representation learning from action-conditioned forecasting of future actions, objects, and motions, enabling encoder reuse across embodiments via decoder-only finetuning;
\textbf{(3)} an empirical characterization of when structured relational inductive biases are beneficial: across eleven bimanual tasks from two datasets, we find that the benefit of our representation over simpler sequence-based baselines grows with task variability in action ordering and object involvement;
\textbf{(4)} a real-robot action planner that couples learned semantic and geometric predictions at deployment, and evidence of its system-level utility: our representations achieve full task success on a bimanual robot across two tasks, outperforming graph ablations, Transformer and decoder-only baselines, and a finetuned VLM under similar data budgets.

\begin{figure*}[h!]
    \centering
    \includegraphics[width=\textwidth]{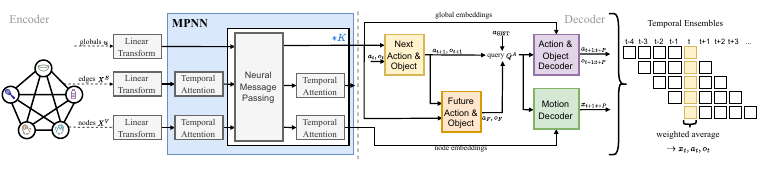}
    \caption{Model architecture: the graph encoder transforms features into embeddings via the MPNN, and the decoders forecast actions, objects, and motions.}
    \label{img:model}
    \vspace{-1em}
\end{figure*}

\section{Related Work}

Learning manipulation task progression from human demonstrations requires representations that capture both semantic task structure and geometric scene evolution. Prior graph-based work focused mainly on skeleton-based representations~\cite{xiaohan2015joint,li2021directed}, some leveraging spatial-temporal GNNs~\cite{li2021symbiotic,tao2021scene}, but skeletal representations are less suited to tabletop manipulation, where relevant structure lies in object-level relations. Scene graphs~\cite{armeni20193d,gay2019visual,kim20193,wald2020learning} provide a natural framework for relational structure and are widely used for Human-Object Interaction~\cite{morais2021learning,wu2025hiergat} --- typically for frame-level action recognition, without modeling task progression.

Closer to our setting, Dreher et al.~\cite{dreher2019learning} use a scene graph of semantic spatial relations between objects and hands for frame-wise action classification and segmentation in bimanual human demonstrations. Their graph network operates at the frame level and does not forecast future actions, object involvement, or object motion. 
Razali et al.~\cite{razali2023action} take the complementary approach, learning fine-grained whole-body motion prediction from a geometric scene graph, action-conditioned but zero-shot from initial object positions without online adaptation. 
Lagamtzis et al.~\cite{lagamtzis2023graph} combine action recognition, future action prediction, and motion forecasting, incorporating geometric information as node features. Their method, however, neglects semantic edge features and global task context, and focuses on single-step unimanual predictions, limiting its reasoning of bimanual multi-step task progression.

Beyond graph-based approaches, manipulation task representations span a broad spectrum, from symbolic formalisms such as PDDL and TAMP~\cite{garrett2021integrated}, requiring manual domain specification, to LLMs/VLMs as general-purpose task backbones~\cite{ahn2022can,liang2023code} or end-to-end VLA policies~\cite{kim2024openvla, black2024pi0}. While attractive in generalisation, the latter rely on large pretraining corpora and extensive finetuning, a requirement at odds with low-data regimes, and their discrete-token backbones do not naturally extend to continuous multi-object motion forecasting. In contrast, object-centric graph representations remain compact and learnable from small demonstration sets, a trade-off we examine empirically in Section~\ref{sec:robot_experiments}.

In contrast, we propose a semantic--geometric task-graph representation that unifies inter-object semantic relations, per-object motion histories, and global task context, and jointly forecasts future actions, associated objects, and object motions from a shared graph embedding. This retains the data efficiency of object-centric graphs while combining symbolic task structure with continuous motion forecasting, and transfers to a physical bimanual robot for online action planning.

\section{Problem Statement} \label{sec:problem}

We consider bimanual tabletop manipulation tasks performed by a human demonstrator and subsequently transferred to a bimanual robot. 
Each task is a sequence of high-level action primitives per hand, each acting on a single target object. 
For \textbf{task structure}, we assume tasks can be segmented into such primitives and each demonstration frame is annotated with a categorical action label and corresponding target object per hand, either manually or via standard action-segmentation pipelines~\cite{dreher2019learning}. 
For \textbf{perception}, a third-person RGB-D camera tracks the 3D object and hand centroids and bounding boxes; semantic inter-object relations (e.g., contact, proximity, direction) are computed from object positions via a heuristic relation taxonomy~\cite{ziaeetabar2018recognition}, requiring no manual annotation. Graph features are precomputed during offline training and computed online at deployment. 
For \textbf{robot deployment}, the bimanual robot is equipped with a library of parameterizable motion primitives - in our instantiation, Probabilistic Movement Primitives (ProMPs)~\cite{paraschos2013probabilistic} - corresponding to the action labels and conditioned on the target object's pose at execution time. We assume access to a small set of robot demonstrations (10-15 per task) for decoder-only finetuning. 
Our representation uses 3D positions rather than 6D poses, which suffices for tasks where the target of each primitive is well-defined and pose-specific manipulation (e.g., in-hand rotation) is not required.

\section{Learning Semantic-Geometric Task Graph-Representations}

We propose an approach for learning semantic-geometric graph-based task representations from bimanual human demonstrations. 
Each task is modeled as a sequence of actions that span multiple time steps. The objective is to predict future task progression by modeling past object movements and semantic relationships in a graphical structure. 
We formalize a manipulation demonstration as a spatio-temporal graph $\mathcal{G}^{(t)} = (\mathcal{V}, \mathcal{E}, u)$ whose node, edge, and global features evolve over $H$ past frames. Using an extended message-passing formulation~\cite{battaglia2018relational}, our encoder jointly learns node embeddings $h_v$, edge embeddings $f_{vw}$, and global embeddings $g_u$ through iterative information exchange, while the decoder forecasts future actions, associated objects, and object motions over a prediction horizon $P$ (Figure~\ref{img:pipeline}).

\subsection{Semantic-Geometric Graph Structure} \label{sec:graph_rep}
Let $\mathcal{D}_{\text{raw}}$ be a dataset of manipulation demonstrations captured via RGB-D video. 
We slice each video into temporal slices with $H$ historical frames and $P$ future frames.
For each frame $t$, we construct a fully-connected, bi-directional spatial-temporal graph that encodes the scene state at frame $t$ along with its history. 
Additionally, we annotate each frame $t$ with the current bimanual actions $a_t = (a^R_t, a^L_t)$ and their corresponding objects $o_t = (o^R_t, o^L_t)$.

\subsubsection{Node Features}
The nodes represent objects in the scene and the user's hands. 
Going forward, \textit{objects} always include the hands.
Unlike related work~\cite{dreher2019learning, lagamtzis2023graph} which create separate nodes per time step in the graph's history, we use a single node per object with temporal features to reduce computational complexity and memory overhead.
The node feature matrix $\bm X^V$ consists of the concatenation of the node's one-hot encoded object ID $c_n$ as well as its 3D coordinates $\bm x_n$ from $H$ past frames, sampled at a rate $S$ (i.e. every $S$-th frame). 
Parameters $H$ and $S$ control the length and temporal granularity of the history.
The complete node feature matrix $\bm X^V_{t-H+1:t} \in \mathbb{R}^{N \times H \times d_V}$ is constructed as
$ \bm X^V[n, t] = [c_n; \bm x_{n, t}] \quad \forall n \in N, \forall t \in \text{history}$, with
$d_V$ as the sum of possible object classes and coordinate dimensionality.

\subsubsection{Edge Features}
Edges capture semantic relationships between nodes, including static spatial relations (e.g., one object is right of the other) and dynamic movements (e.g., one object approaches another object)~\cite{ziaeetabar2018recognition}. Multiple spatial relations can hold true for a pair of objects. The edge feature matrix $\bm X^{E}_{t-H+1:t} \in \mathbb{R}^{M \times H \times d_E}$ is defined over the last $H$ time steps sampled at a rate $S$.
We can therefore use a multi-hot encoding to represent the relations between two nodes, where $d_E$ is the number of possible relations. 

\subsubsection{Global Features}
We use a global variable $u$ to encode information relevant to the entire graph and not specific to a certain node or edge.
In our case, this global variable consists of an one-hot encoded task ID of the current task. 
In contrast to node and edge features, the global features $u \in \mathbb{R}^{d_U}$ are not defined over the history of the graph but just the current time step, with $d_U$ being the number of possible tasks.

\subsection{Graph Neural Network Architecture} \label{sec:model}
Our goal is to predict future actions, action-objects, and object coordinates over the prediction horizon $P$. Our architecture follows an encoder-decoder structure: the graph is first encoded through an MPNN that learns graph embeddings by propagating information across the graph, and these embeddings are then passed through multiple prediction heads to forecast the future progression of the task.
The overall architecture is depicted in Figure \ref{img:model}.

\subsubsection{MPNN Encoder}
Node, edge, and global features are encoded separately via linear transformations into a shared hidden dimension $d_{\text{MP}}$. 
Global embeddings are tiled to share the same temporal dimension $H$ as node and edge embeddings.
To consider the temporal aspect of the task, all embeddings are encoded with Rotary Position Embeddings (RoPE)~\cite{su2024roformer} over their temporal dimension.
The frame id $t$ corresponding to each entry in the embedding is used to choose the rotation parameters for that embedding.

Subsequently, the MPNN refines node, edge, and global embeddings iteratively over $K$ iterations. 
In each iteration, first the edge embeddings are updated, followed by the node embeddings and the global embeddings.
The edge embeddings are updated using linear transformations with learnable weights $\bm W^E_k$ and non-linearities $\alpha$:
\begin{small}
\begin{align*}
    f_{vw}^{(k+1)} 
    &\mathrel{+}= 
    \alpha \Big(
        \bm{W}^E_{1,k}
        \big[
            \alpha\big(\bm{W}^E_{2,k} f_{vw}^{(k)}\big);
            \alpha\big(\bm{W}^E_{3,k} [h_v^{(k)}; h_w^{(k)}]\big); 
            g_u^{(k)}
        \big]
    \Big).
\end{align*}
\end{small}
Compared to~\cite{dreher2019learning}, we use individual weights $\bm W^E_k$ per message passing iteration $k$, allowing to learn different features per iteration, and add residual connections to refine features across iterations.
Node and global updates follow similarly.
We alternate message passing with temporal self-attention over the temporal dimension of node/edge features. 
This helps the model learn dependencies across time steps.

\subsubsection{Prediction Decoder Models}
The decoder predicts three components: future actions, associated objects, and object motions, as object-level 3D trajectories over horizon
$P$, via a Transformer decoder attending to node embeddings. Since high-level actions and low-level motions operate at different time scales (actions span multiple steps while motions change frame-by-frame), we design a multi-stage decoder to handle these frequencies.
In the first stage, an MLP predicts the immediate next action-object pair $a_{t+1}, o_{t+1}$ from the current pair $a_t, o_t$ and temporally averaged global graph embeddings $\bar{g}_u^K$, capturing immediate task progress.

In the second stage, we predict future semantic action-object pairs $a_F, o_F$, i.e. the next high-level action after the current one terminates, using an MLP conditioned on the predicted pair $a_{t+1}, o_{t+1}$ and temporally averaged global embeddings $\bar{g}_u^K$. 
This decouples high-level action forecasting from the current action's duration, enabling independent reasoning over task sequencing. Separating scene representation learning in the encoder from action-conditioned reasoning in the decoder prevents entangling action labels with perceptual representations while explicitly reasoning over task progress.

The overall model should predict actions, objects, and motions for each time step in the horizon $P$. We construct action-object queries that capture both action-object context and temporal information. Task progression depends on motion history (encoded in graph embeddings) and semantic high-level action-object history. We track the $n_{\text{past}}$ most recent action-object pairs per hand, as action switches occur at different times for left and right hands. 
For notational brevity, we denote action-object pairs as $a_i = (a_i, o_i)$.
For each hand $i \in \{R, L\}$, we build a query $a_{i, Q} = [a_{i,\text{HIST}} | a_{i, t+1} | a_{i, F}]$, concatenating the past pairs $a_{i,\text{HIST}}$, predicted next pairs $a_{i, t+1}$, and predicted future semantic pairs $a_{i, F}$.

Action queries from both hands are concatenated and linearly projected into embedding space, with temporal structure encoded via RoPE embeddings~\cite{su2024roformer} based on action start frames, forming the final query vector $Q^A$. This feeds two separate Transformer decoders~\cite{vaswani2017attention}: the action-object decoder attends to the (non-averaged) global embeddings $g_u^K$ to predict action and object sequences, while the motion decoder attends to node embeddings $h_v^K$ to predict motion sequences. Both outputs are projected to their output spaces via linear layers.

\subsection{Training for Multi-Step Task Prediction} \label{sec:training}

We train with a joint loss combining classification and regression objectives. 
Action and object classifiers use weighted cross-entropy (CE) to compensate for class imbalances, while motion prediction uses mean squared error (MSE) between predicted and ground-truth coordinates.
The total loss function is a weighted sum of these losses
\begin{align*}
\mathcal{L}^{\text{Total}} = \sum_a \mathcal{L}^{\text{CE}}_a + \sum_o \mathcal{L}^{\text{CE}}_o + \beta_{\text{MSE}}\, \mathcal{L}^{\text{MSE}}_{x_{t+1:t+P}},
\end{align*}
where action losses include $\{a_{t+1}, a_F, a_{t+1:t+P}\}$, object losses include $\{o_{t+1}, o_F, o_{t+1:t+P}\}$ and $\beta_{\text{MSE}}$ balances regression and classification losses. 
AdamW is used as optimizer.

The auxiliary next and future action-object predictions ($a_{t+1}, o_{t+1}$ and $a_F, o_F$) are used to guide the model towards decoupling high-level action forecasting from the action's duration, but do not enable multi-step task execution beyond a fixed prediction horizon $P$ on their own.
To determine which action to execute at each step, we employ \textit{action chunking} \cite{zhao2023learning}.
For each time step $t$, the predicted actions $a_{t+1:t+P}$, objects $o_{t+1:t+P}$, and motions $\bm x_{t+1:t+P}$ produce overlapping predictions, that are reconciled using temporal ensembles with exponential decay weighting, where higher weights are assigned to older predictions.
Final predictions are computed as weighted averages of all overlapping predictions.

Following this training regime, our model integrates object relationships and spatial-temporal reasoning to predict future task progression, including both short-term and long-term action and motion predictions. 
Leveraging action chunking and temporal ensembles ensures smooth, continuous predictions across overlapping prediction windows.

\subsection{Coupled Online Action Planning} \label{sec:planning}

At deployment, the trained model is used within an online action planning framework that couples its semantic and geometric outputs into a single decision. High-level actions are executed on the robot via a library of parameterizable ProMPs~\cite{paraschos2013probabilistic}, one per action type. 
Unlike the learned task model operating on 3D object positions, the ProMPs are learned over full 6D task-space poses together with gripper openings from teleoperated demonstrations. 

After executing a primitive, the planner selects the next action-object pair as follows. 
The decoder outputs $a_{t+1:t+P}$ and $o_{t+1:t+P}$ are processed via action chunking (see Section~\ref{sec:training}), yielding action and object logits treated as categorical probabilities $p(a)$ and $p(o)$. 
A symbolic precondition mask $M(s_t) \in \{0, 1\}^{|\mathcal{A}|\times|\mathcal{O}|}$ zeros out infeasible action-object pairs in the current state $s_t$ (e.g., grasping an already-held object). 
This results in a joint action-object probability $p(a, o) \propto p(a)p(o)M(s)$.
For the top-$k$ remaining candidates, a trajectory score $s^{\text{Traj}}(a, o)$ is computed: the model's motion predictions are bootstrapped over a short rollout horizon $T$ conditioned on the action-object candidate and the corresponding ProMP is conditioned on the current state. A Mahalanobis distance score between the predicted rollout $\bm x$ and the ProMP mean trajectory $\bm \mu$ and covariance $\Sigma$ is computed
\begin{align*}
    s^{\text{Traj}}(a, o) &= \exp\Big(- \frac{1}{3T} \sum^T_{t=1} (\bm x_t - \bm \mu_t)^T \Sigma_t^{-1} (\bm x_t - \bm \mu_t)\Big).
\end{align*}
An additional exponential penalty is applied to rollouts that deviate substantially from the current state.
This results in a final score for an action primitive $s(a,o) = p(a,o) \cdot s^{\text{Traj}}(a, o)$.
Before the highest-scoring candidate is executed, a physical feasibility check ensures that candidate remains within the robot's workspace; if not, the planner is re-queried.

This scoring couples the decoder's two outputs at decision time: the action-object prediction determines \textit{what} to do, while the motion prediction verifies consistency with the learned primitives. Candidates with plausible symbolic predictions but inconsistent motion are down-weighted, and vice versa. Left- and right-hand actions are scored independently, as the bimanual action boundaries are asynchronous.

\section{Experiments and Results}\label{sec:robot_experiments}

\begin{figure}[t]
    \centering
    \includegraphics[width=0.46\textwidth]{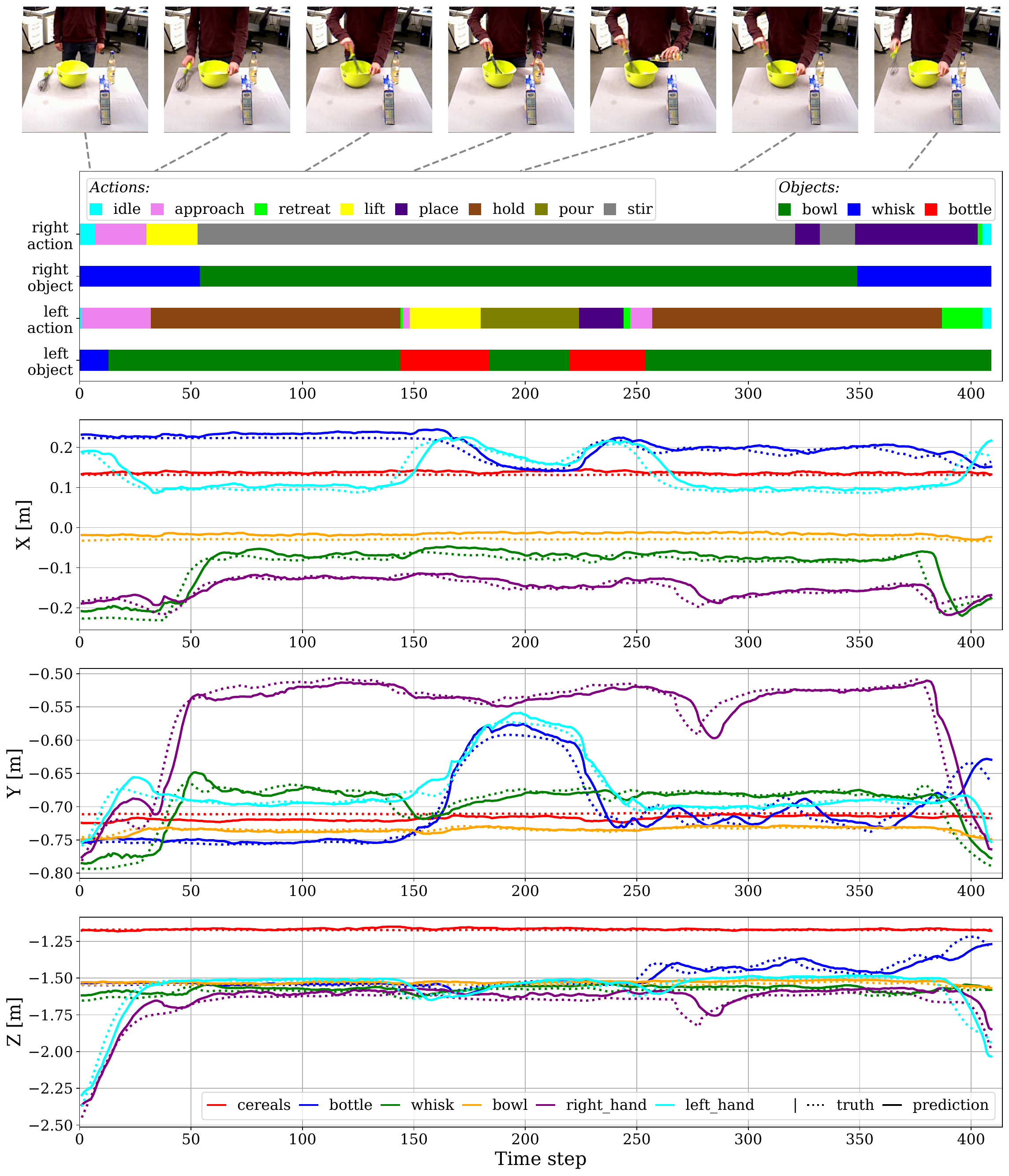}
    \caption{A qualitative example from KIT(Bimacs) of the \textit{cooking} task. 
    Key frames (top row) are aligned with the action and object predictions of our MPNN model over time. 
    The 3D motion predictions are shown below. Solid lines denote model predictions and dashed lines denote the ground-truth.
    }
    \vspace{-1em}
    \label{fig:predictions}
\end{figure}

In this section, we evaluate whether semantic–geometric graph representations learned from human demonstrations capture long-horizon task structure and generalize across tasks, subjects, and embodiments. Our experiments are designed to assess (i) when structured relational inductive biases are beneficial, and (ii) whether the learned representations can be transferred to a physical bimanual system. We evaluate the models on the KIT Bimanual Actions Dataset (Bimacs)~\cite{dreher2019learning}, containing RGB-D recordings of bimanual demonstrations for five cooking and four workshop tasks by six subjects with ten takes each. 
The dataset provides ground-truth hand actions, 3D object bounding boxes, and semantic object–object relations. 
We further annotated action–object labels, removed object detections in the background, slightly smoothed the trajectories, standardized trajectories across the training set, and augmented the data via mirroring and temporal resampling.

To adapt the models to our setting, we collected additional data of four tabletop tasks: the \textit{cooking} and \textit{wiping} tasks from KIT(Bimacs), plus two new tasks —  clearing up objects on the table, and taking out objects from a box. Each task was performed 10 times by four subjects, denoted as Ours(Bimacs). 

\subsection{Baseline Architectures}

To show the benefits of our GNN-based encoder, we compare the performance of our encoder with four different encoder architectures. The following baselines are used to probe the role of different inductive biases in learning task representations.
Each model uses a different encoder structure, but uses our decoder to ensure comparability between the models for our specific prediction objectives.

\textbf{Dreher~\cite{dreher2019learning}} An iterative MPNN that encodes scene structure using object IDs as node features and semantic relations as edge features, ignoring geometric information. GNN weights are shared across 10 message passing iterations.

\textbf{Lagamtzis~\cite{lagamtzis2023graph}} A deep relational graph convolutional network (RGCN) that updates only node embeddings, neglecting edge and global features. Node features include object identifiers and geometric information. We use 20 RGCN blocks (vs. 36 in the original) for parameter comparability.

\textbf{Transformer} A classical transformer encoder~\cite{vaswani2017attention} operating on sequential node features (object ids and geometric information), without the graph structure or semantic edge features. Uses 8 encoder layers with 16 attention heads each.

\textbf{Decoder-Only} The encoder is omitted entirely; node features are projected into the embedding space and fed to the decoder. This baseline quantifies the encoder's contribution by removing the message passing and structured context.

\textbf{MPNN (Ours)} The GNN-encoder described in Section~\ref{sec:model}, with $K=3$ message passing iterations, 2 temporal attention heads, and Leaky-ReLU activations.

An overview of all models is shown in Table \ref{tab:model_comparison}. Dreher~\cite{dreher2019learning} and Lagamtzis~\cite{lagamtzis2023graph} serve as GNN-ablations, isolating the contributions of combining semantic edge features with geometric node features. 
The Transformer and Decoder-Only models are included as non-graph baselines to assess the benefits of the graph structure.
The hyperparameters of all encoder variants were selected such that their total parameter counts fall within the same range as our proposed encoder 
(MPNN 449K parameters, Transformer 418K, Decoder-Only 216K, Lagamtzis 462K, Dreher 313K).
Following extensive hyperparameter search, all models use $H=S=P=10$, $d_{MP}=64$, 2-layer Leaky-ReLU MLPs for action/object classifiers, 2-layer transformer decoders with 4 attention heads, $n_\text{past}=20$, batch size 128, and $\beta_\text{MSE}=1000$.
Models were trained on an HPC cluster in under one day per task.

\begin{table}[t]
\centering
\caption{Overview over the different baseline models in this work.}
\resizebox{0.49\textwidth}{!}{
\begin{tabular}{|l|c|c|c|c|c|}
\cline{2-6}
\multicolumn{1}{l|}{} & Dreher & Lagamtzis & Trans- & Decoder- & MPNN \\
\multicolumn{1}{l|}{}  & \cite{dreher2019learning} & \cite{lagamtzis2023graph} & former & Only & (Ours) \\
\hline
Graph-based & \checkmark & \checkmark &  &  & \checkmark \\
\hline
Geometric info &  & \checkmark & \checkmark & \checkmark & \checkmark \\
\hline
Semantic info & \checkmark &  &  &  & \checkmark \\
\hline
Encoder-based & \checkmark & \checkmark & \checkmark &  & \checkmark \\
\hline
\end{tabular}
}
\label{tab:model_comparison}
\vspace{-1em}
\end{table}

\begin{figure*}[t]
    \centering
    \includegraphics[width=0.99\textwidth]{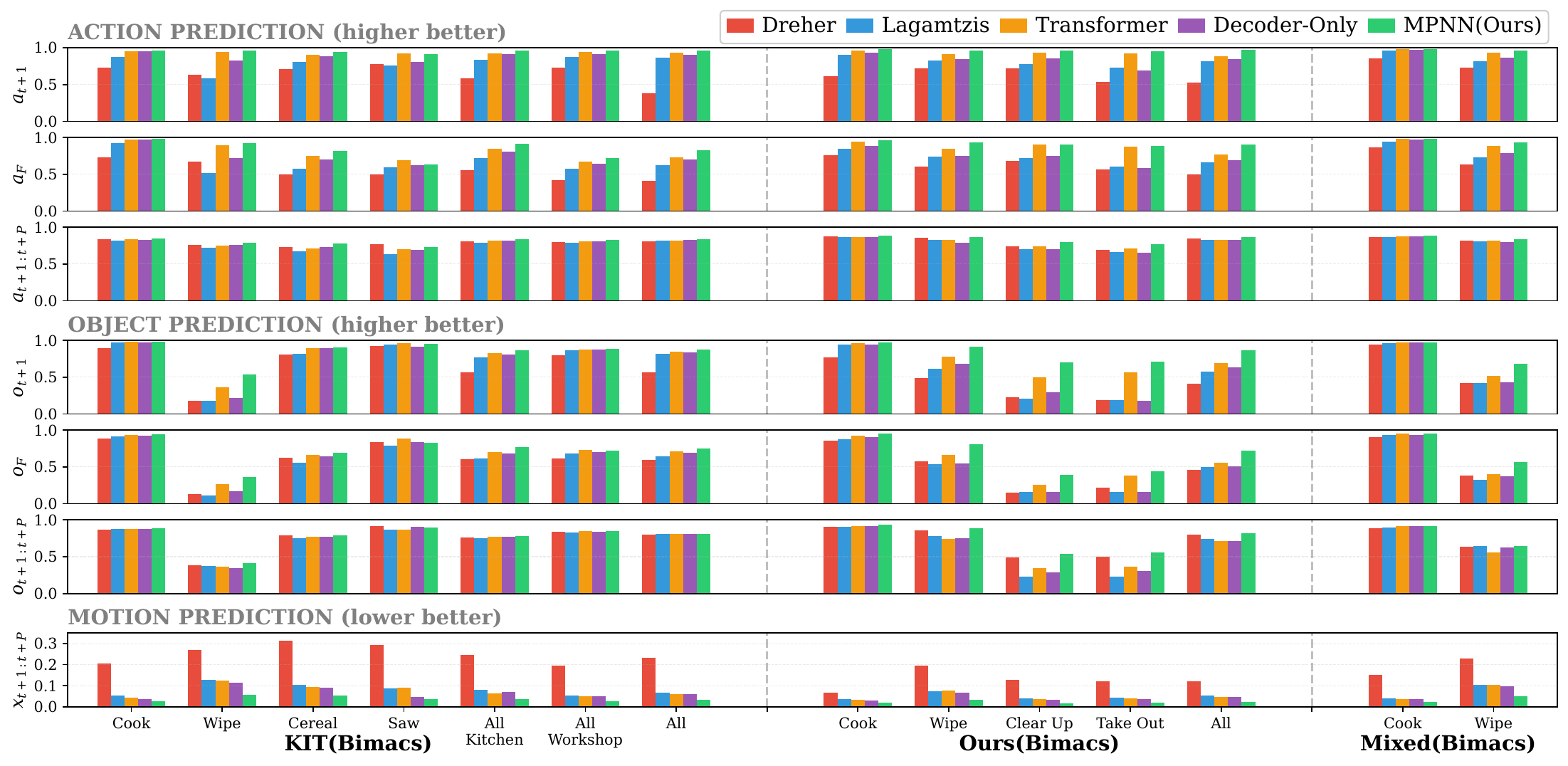}
    \caption{
    Cross validation results on selected tasks from the KIT(Bimacs), from Ours(Bimacs) and a mixed dataset, including also multi-task models.
    Action prediction accuracies (next $a_{t+1}$, future $a_F$ and horizon $a_{t+1:t+P}$) are shown on top, object predictions in the middle (next $o_{t+1}$, future $o_F$ and horizon $o_{t+1:t+P}$) and motion prediction ($\bm x_{t+1:t+P}$) RMSE in the bottom row.
    The baselines are shown in different colors. 
    }
    \vspace{-1em}
    \label{fig:bar_results_act_obj}
\end{figure*}

\subsection{Prediction Results on Human Demonstrations}
We begin by examining results from training models on human demonstrations. All models are trained separately on each task from the two human datasets and in multi-task settings. Figure \ref{fig:predictions} visualizes our MPNN model's performance on an exemplary \textit{cooking} task demonstration. The task involves one hand approaching, lifting, and stirring with a whisk before placing it down, while the other hand holds the bowl, pours from a bottle, and holds the bowl again. The plot shows motion predictions for each object, plus action and object predictions for both hands. Motion predictions are accurate and match ground truth trajectories. Action predictions contain all relevant actions and align with object predictions, however, certain reaching actions (approach and retreat) are predicted for short time spans that do not reflect true execution time.
In the end, right hand predictions flicker between place and stir.

Figure \ref{fig:bar_results_act_obj} shows results for action, object, and motion predictions across tasks from KIT(Bimacs) and Ours(Bimacs) and a mixed version of both datasets. Evaluation uses leave-one-subject-out cross-validation over human subjects, averaging over 4 seeds per setting. Model performance is compared using classification accuracy for action and object predictions, and RMSE for motion prediction.
During training, we observed that MPNN models converge faster than the other models. We evaluate all models at that same epoch to provide comparable training data across models.

\textbf{KIT(Bimacs)}
We begin by analyzing single-task results on KIT(Bimacs). For the auxiliary semantic next and future action predictions, MPNN and the Transformer achieve the highest accuracies, indicating their ability for accurate task progression prediction. The Decoder-only model achieves similar accuracies on simple tasks like \textit{cooking}, but underperforms on complex tasks with higher object variation like \textit{wiping} or action variation like \textit{cereals}, revealing limitations of predicting task progression solely through memorizing past action/object sequences. Dreher~\cite{dreher2019learning} and Lagamtzis~\cite{lagamtzis2023graph} perform poorly on semantic actions, demonstrating the need of combining geometric and semantic scene information. 

For action prediction over horizon $P$, model differences are smaller compared to semantic action predictions, though MPNN consistently benefits from modeling semantic–geometric relations, particularly in tasks with high action or object variability. This objective requires predicting the next $P$ actions independently of individual action durations; models achieve high accuracies by simply repeating the current action without considering past semantic-geometric information or learning future task progression. Conversely, Dreher~\cite{dreher2019learning} performs better on horizon action prediction than Lagamtzis~\cite{lagamtzis2023graph}, while for semantic action predictions it is the other way around. Dreher~\cite{dreher2019learning} outperforms MPNN on  horizon predictions of \textit{sawing}, but falls back for the semantic ones.

\begin{table*}[h]

\centering
\caption{Cross-Validation Results of training the MPNN (Ours) Model on robot demonstrations, evaluating mixed models trained on human demonstrations on the robot dataset, and finetuning those models on robot data.}

\resizebox{\textwidth}{!}{
\begin{tabular}{c| l || c | c | c | c | c | c | c |}
\cline{2-9}
& \multirow{2}{*}{Method} & \multicolumn{6}{c|}{Accuracy $(\uparrow)$} & RMSE [m] \\
\cline{3-8}
& & $a_{t+1}$ & $a_F$ & $a_{t+1:t+P}$ & $o_{t+1}$ & $o_F$ & $o_{t+1:t+P}$ & $(\downarrow)$ \\
\hline

 \multicolumn{1}{|c|}{\multirow{3}{*}{\rotatebox[origin=c]{90}{\scriptsize Cooking}}}  & Robot & 0.7460$\pm$0.0501 & 0.6594$\pm$0.0362 & 0.6443$\pm$0.0453 & 0.8452$\pm$0.0317 & 0.8538$\pm$0.0087 & 0.7840$\pm$0.0148 & 0.0547$\pm$0.0013 \\
\cline{2-9}

\multicolumn{1}{|c|}{} & Not Finetuned  & 0.9279$\pm$0.0235 & 0.9253$\pm$0.0075 & 0.7661$\pm$0.0108 & 0.8044$\pm$0.0106 & 0.9185$\pm$0.0057 &  0.7950$\pm$0.0062 & 0.0304$\pm$0.0002 \\
\cline{2-9}

\multicolumn{1}{|c|}{}& Finetuned   & \textbf{0.9454$\pm$0.0249} & \textbf{0.9635$\pm$0.0165} & \textbf{0.8815$\pm$0.0166} & \textbf{0.8878$\pm$0.0107} & \textbf{0.9459$\pm$0.0146} & \textbf{0.8664$\pm$0.0059} & \textbf{0.0212$\pm$0.0009} \\
\hline

\multicolumn{1}{|c|}{\multirow{3}{*}{\rotatebox[origin=c]{90}{\scriptsize Clear Up}}} & Robot & 0.9213$\pm$0.0554 & 0.8610$\pm$0.0889 & 0.6784$\pm$0.0797 & 0.5358$\pm$0.0792 & 0.2830$\pm$0.0596 & 0.2052$\pm$0.0612 & 0.0352$\pm$0.0088 \\
\cline{2-9}

\multicolumn{1}{|c|}{} & Not Finetuned  & 0.9515$\pm$0.0091 & 0.8873$\pm$0.0188 & 0.7434$\pm$0.0210 & 0.6671$\pm$0.0401 & \textbf{0.4508$\pm$0.0232} &  0.4870$\pm$0.0432 & 0.0352$\pm$0.0021 \\
\cline{2-9}

\multicolumn{1}{|c|}{} & Finetuned   & \textbf{0.9636$\pm$0.0105} & \textbf{0.9407$\pm$0.0213} & \textbf{0.8347$\pm$0.0231} & \textbf{0.6922$\pm$0.0367} & 0.4278$\pm$0.0282 & \textbf{0.5573$\pm$0.0197} & \textbf{0.0235$\pm$0.0010} \\
\hline

\end{tabular}
}
\label{tab:finetuning}
\end{table*}

Accurate motion forecasting requires integrating past object interactions and relative movements, which are naturally captured by message passing over semantic–geometric task graphs. Object predictions largely follow action predictions, but accuracy differences between models are smaller for simple tasks since multiple subsequent actions often involve the same objects, reducing complexity. For the \textit{wiping} task with more object variety, differences become apparent: our MPNN clearly outperforms all models including the Transformer. Overall object prediction accuracies are lower compared to other tasks, as sub-task order may vary across demonstrations, eliminating unique solutions for object predictions. 

Across all tasks, our MPNN achieves the best motion prediction results, demonstrating its ability to utilize past scene information. Notably, Dreher~\cite{dreher2019learning} underperforms here due to lacking geometric information.

\textbf{Ours(Bimacs)}
Results for \textit{cooking} and \textit{wiping} on Ours(Bimacs) match those on KIT(Bimacs). The tasks (\textit{clear up} and \textit{take out}) were chosen to include high object variation, demonstrating the benefits of our MPNN model on reasoning over past object motions, where the Decoder-only model underperforms. Our MPNN model achieves the highest object prediction accuracy, followed by the Transformer. For object prediction over a horizon $P$, Dreher~\cite{dreher2019learning} ranks second, confirming our KIT(Bimacs) findings that GNN encoders without geometric information predict immediate task progression, but lag in semantic predictions.

\textbf{Multi-task Models}
We also trained multi-task models on all cooking, all workshop, and all tasks of KIT(Bimacs) and Ours(Bimacs). 
The results match single-task observations and
demonstrate that our MPNN encoder learns common concepts and can generalize between different tasks.
Notably, Dreher~\cite{dreher2019learning} performs worse on semantic action and object predictions for multi-task versus single-task models.

\textbf{Mixed(Bimacs)}
The KIT(Bimacs) and Ours(Bimacs) datasets contain demonstrations of two common tasks: \textit{cooking} and \textit{wiping}. Our additional demonstrations recorded in our own setup bridge the gap between human and robotic setups. To investigate whether our model can learn task progression independently of the underlying environment, we mix the two datasets and train all models on Mixed(Bimacs). Results show that our MPNN model's action prediction accuracies do not dramatically decrease on the mixed dataset compared to individual datasets, and even improve for most objectives. The other models also improve when mixing datasets, likely due to the additional training data.

\subsection{Transferring Learned Representations for Online Robotic Bimanual Action Planning}

We evaluate the transfer of our learned task graph-representations to a bimanual robot within an online action planning framework and using a library of ProMPs~\cite{paraschos2013probabilistic} (see Section \ref{sec:planning}). 
For real-robot experiments, we select the \textit{cooking} task from KIT(Bimacs), and the \textit{clear up} task from Ours(Bimacs) dataset, as the object variability enables assessing the benefits of structured representations. 

We compare the direct evaluation of our models (MPNN) trained on the mixed human dataset (or in the \textit{clear up} task, only on Ours(Bimacs)), and finetuning on robot demos (Table \ref{tab:finetuning}).
For the \textit{cooking} task, the 10 robot demos were collected (4 train/ 2 val/ 4 test) by executing ProMPs in a predefined sequence. For the \textit{clear up} task, 15 robot demos (8/3/4) were used with different object combinations.
We compare against models trained only on robot data.
For finetuning, we update only decoder parameters. 
Models trained on limited robot data perform poorly for both tasks, while those pretrained on human demos achieve higher accuracies and lower RMSE, indicating an effective transfer. 
Best performance comes after finetuning, attributed to human-robot motion differences and embodiment mismatch.
For the \textit{clear up} task, we observe significantly higher object accuracy using the human models compared to robot-only training, and even more for finetuned models. This shows the benefit of the object variations seen during human pretraining.
Hence, robot experiments use the finetuned model.

\textbf{Online Action Planning:}
We evaluate the real-world performance of the models by integrating them into the online action planning framework introduced in Section \ref{sec:planning} with $k=4$.
As an additional baseline for real-robot evaluation, we compare against the VLM Florence-2-Base~\cite{xiao2024florence} (0.23B param) finetuned on the same set of robot demonstrations.
The model receives the current RGB image, action-object history, task description, object list, and current object positions as input, and predicts the next action, objects, and next object positions.

\begin{table}[t]

\centering
\caption{Results of online action planning on the real-robot \textit{cooking} and \textit{clear up} tasks with finetuned models. The table shows the overall success rates, the planner infeasibility rates, and motion prediction RMSE.}

\resizebox{0.498\textwidth}{!}{
\begin{tabular}{c| l || c | c | c |}
\cline{2-5}
\small
& Method & Success Rate $(\uparrow)$ & Infeasibility Rate $(\downarrow)$ &  RMSE [m] $(\downarrow)$ \\

\hline

\multicolumn{1}{|c|}{\multirow{6}{*}{\rotatebox[origin=c]{90}{Cooking}}} & Dreher \cite{dreher2019learning} & \textbf{1.0000$\pm$0.0000} & 0.0429$\pm$0.1020 & 0.1569$\pm$0.0229 \\

\multicolumn{1}{|c|}{} & Lagamtzis \cite{lagamtzis2023graph} & 0.7250$\pm$0.1750 & 0.0890$\pm$0.1741 & 0.0854$\pm$0.0036 \\

\multicolumn{1}{|c|}{} & Transformer & \textbf{1.0000$\pm$0.0000} & \textbf{0.0000$\pm$0.0000} & 0.0509$\pm$0.0043 \\

\multicolumn{1}{|c|}{} & Decoder-Only & \textbf{1.0000$\pm$0.0000} & \textbf{0.0000$\pm$0.0000} & 0.0544$\pm$0.0022  \\

\multicolumn{1}{|c|}{} & VLM & 0.0000$\pm$0.0000 & 1.0000$\pm$0.0000 & 88.1708$\pm$8.4930  \\

\multicolumn{1}{|c|}{} & MPNN (Ours) & \textbf{1.0000$\pm$0.0000} & \textbf{0.0000$\pm$0.0000} & \textbf{0.0469$\pm$0.0022} \\

\hline

\multicolumn{1}{|c|}{\multirow{6}{*}{\rotatebox[origin=c]{90}{Clear Up}}} & Dreher \cite{dreher2019learning} & 0.7990$\pm$0.2481 & 0.4449$\pm$0.2206 &  0.1681$\pm$0.0236 \\

\multicolumn{1}{|c|}{} & Lagamtzis \cite{lagamtzis2023graph} & 0.3320$\pm$0.3430 & 0.6605$\pm$0.2604 & 0.1873$\pm$0.0169 \\

\multicolumn{1}{|c|}{} & Transformer & 0.7320$\pm$0.2265 & 0.3899$\pm$0.2673 & 0.0622$\pm$0.0058 \\

\multicolumn{1}{|c|}{} & Decoder-Only & 0.7390$\pm$0.2165 & 0.3298$\pm$0.3184 & 0.0540$\pm$0.0046  \\

\multicolumn{1}{|c|}{} & VLM & 0.0000$\pm$0.0000 & 1.0000$\pm$0.0000 & 18.3042$\pm$14.5292  \\

\multicolumn{1}{|c|}{} & MPNN (Ours) & \textbf{1.0000$\pm$0.0000} & \textbf{0.1079$\pm$0.1353} & \textbf{0.0440$\pm$0.0039} \\

\hline
\end{tabular}
}
\label{tab:planning}
\end{table}

Table~\ref{tab:planning} reports results over 10 trials per baseline on the \textit{cooking} task (varied object placements) and \textit{clear up} task (varying object types). Task success measures the rate of completed sub-tasks; infeasibility rate the fraction of planner queries where the top-scored candidate passes the symbolic precondition mask but is rejected by the downstream physical feasibility check (e.g., hand-side mismatches, missing primitives, workspace violations), triggering automatic requery. A trial fails only when the same query is rejected 10 times consecutively. We also report motion prediction RMSE against tracked object positions.

For \textit{cooking}, all models except Lagamtzis~\cite{lagamtzis2023graph} achieve full task success and low infeasibility rates, attributable to low task complexity — the action-object sequence is fixed across demonstrations — while our model achieves the lowest RMSE. For the more complex \textit{clear up} task, only our model achieves full task success, the lowest infeasibility rate, and lowest motion RMSE, demonstrating clear benefits from structured representations under high object variability. The transformer and decoder-only baselines show higher infeasibility rates due to unsafe predictions (e.g. approaching objects on opposite sides of the table). 
The VLM achieves zero success on both tasks, failing to switch from the currently executing action and producing very high motion RMSE given its text-based nature.

\subsection{Discussion}
Results on human demonstrations indicate that semantic–geometric task graph-representations support more accurate modeling of long-horizon task progression, particularly for motion forecasting and object-centric reasoning. The MPNN encoder consistently benefits from jointly encoding semantic relations and geometric evolution, highlighting the importance of integrating both information modalities into the graph. In contrast, encoders with weaker relational inductive biases struggle to capture task progression in demonstrations with high action and object variability, where sequence-level reasoning alone is insufficient.

These observations suggest that task variability — both in terms of action ordering and object involvement — is a key factor in determining when structured relational representations are advantageous. When tasks exhibit limited variation, simpler architectures can suffice; however, as the complexity of the interaction increases, explicitly modeling semantic–geometric relations becomes increasingly important for learning transferable task abstractions.

These results are enforced when transferring the learned models to a bimanual robot and integrating both the model's action-object and motion predictions into an online action planning framework. 
Learned task representations successfully transfer to a physical bimanual robot, but require finetuning on robot demonstrations for successful task execution.
The real-robot experiments show that particularly for the \textit{clear up} task with high object variability, our MPNN model is the only model able to result in complete task success.

\section{Conclusion and Outlook}

In this paper, we introduced a semantic–geometric task graph-representation for bimanual manipulation that jointly encodes object identities, inter-object semantic relations, and object motion histories within a unified graph structure. A decoupled GNN encoder-decoder architecture separates scene representation learning from action-conditioned forecasting, enabling encoder reuse across embodiments. We demonstrated that the learned task representations from human demonstrations successfully transfer to a real bimanual robot to drive online action planning. At deployment, an online action planner couples the decoder's semantic action-object predictions with geometric motion predictions scored against learned Probabilistic Movement Primitives. When compared against other state-of-the-art GNN-based approaches~\cite{dreher2019learning,lagamtzis2023graph},  Transformers, Decoder-only models, and a finetuned VLM, our results across eleven bimanual tasks highlight that the benefit of structured semantic–geometric representations over simpler sequence-based models, especially with high action-object variability, is involved in a task.

Our approach has several limitations that point to concrete future directions. While 3D object positions suffice for the tasks considered, incorporating 6D poses into the graph representations~\cite{brandstetter2022geometric} would extend applicability to orientation-critical manipulation. 
We will also explore using a VLM as a perceptual front-end to ground graph features in rich visual representations. Complementarily, the scene graphs produced by our approach yield a structured pretraining resource for spatially-grounded VLMs, requiring only 3D object tracks and heuristic semantic relations, endowing VLMs with stronger spatial and causal priors over object interactions toward scalable and data-efficient representations. Moreover, motion predictions are currently used only for action scoring. A natural extension is to couple the structured semantic-geometric predictions with VLAs as a high-level reasoning substrate that guides and grounds the low-level generative policies of VLAs toward more interpretable and data-efficient manipulation.

\addtolength{\textheight}{-12cm}   




\section*{Acknowledgement}
This research is funded by the EU Horizon Europe Projects \enquote{MANiBOT} (101120823), \enquote{ARISE} (101135959), the German Research Foundation (DFG) Emmy Noether Programme (CH 2676/1-1), the German Federal Ministry of Research, Technology and Space (BMFTR) Project \enquote{RIG} (16ME1001), and the German Federal Ministry of Research, Technology and Space (BMFTR) Project \enquote{IKIDA} (01IS20045). 
This work is supported by the European Research Council (ERC) grant \enquote{SIREN} (101163933), funded by the European Union. Views and opinions expressed are however those of the author(s) only and do not necessarily reflect those of the European Union or the European Research Council Executive Agency. Neither the European Union nor the granting authority can be held responsible for them.

The authors gratefully acknowledge the computing time provided to 
them on the high-performance computer Lichtenberg II at TU Darmstadt, 
funded by the German Federal Ministry of Research, Technology and Space (BMFTR)
and the State of Hesse.


\bibliographystyle{IEEEtran}
\bibliography{references}

\end{document}